\documentclass[journal]{IEEEtran}
\ifCLASSINFOpdf
\else
\fi
\usepackage{algorithm}
\usepackage{algorithmic}
  
\usepackage{amsmath}
\usepackage{newtxtext}
\usepackage{newtxmath}

\usepackage{amsmath}
\usepackage{amssymb}
\usepackage{tabularx}
\usepackage[noadjust]{cite}
\usepackage{booktabs}
\usepackage[utf8]{inputenc}
\usepackage{booktabs} 
\usepackage{multirow}
\usepackage{orcidlink}
\usepackage{enumitem}
\usepackage{graphicx}

\begin{document}
\title{CMPF:Cross-modal Progressive Filtering for Referring Remote Sensing Image Segmentation}
%
%
%

\author{Yan~Zhang, %
        \and LiangKe~Gao, %
        \and RunFeng~Luo, %
        \and Lin~Yuan, %
        \and Boyuan~Li, %
        \and Xinbo~Gao,~\IEEEmembership{IEEE~Fellow}%
       \thanks{Yan Zhang (e-mail: yanzhang1991@cqupt.edu.cn), LiangKe Gao (e-mail: s240231188@stu.cqupt.edu.cn), RunFeng Luo (e-mail: s240201160@stu.cqupt.edu.cn), and Lin Yuan (e-mail: yuanlin@cqupt.edu.cn) are with the School of Computer Science and Technology, Chongqing University of Posts and Telecommunications, Chongqing 400065, China. Yan Zhang is also with the State Key Laboratory of Integrated Service Networks. Boyuan Li is with the China North Vehicle Research Institute, Beijing, China. Xinbo Gao is with the School of Artificial Intelligence, Xidian University, Xi'an 710071, China, and the corresponding author (e-mail: gaoxb@xidian.edu.cn).}
}

%
%

\markboth{Journal of \LaTeX\ Class Files,~Vol.~13, No.~9, September~2026}%
{Shell \MakeLowercase{\textit{et al.}}: Bare Demo of IEEEtran.cls for Journals}
%



\maketitle


\begin{abstract}
    Referring Remote Sensing Image Segmentation (RRSIS) aims to segment specific regions in remote sensing images guided by natural-language referring expressions. Although substantial progress has been made in improving vision–language alignment, multimodal information propagation across scales remains underexplored. Most existing methods independently construct language-aware visual representations at each scale and propagate them in a coarse-to-fine manner, but they do not explicitly distinguish which information should be transferred; consequently, expression-relevant responses and irrelevant activations are transmitted together and progressively accumulated across scales, impairing target localization in cluttered remote sensing scenes. To address this issue, we propose a cross-modality progressive filtering network (CMPF). At each scale, the CMA block aligns the referring expression with the current visual feature; the RF block then produces a referential response map that suppresses the regions inconsistent with the expression, and the filtered feature is injected into the next scale through a complementary gate, where it is aligned and filtered again. As a result, the cross-scale propagation is transformed from a one-off fusion into a progressive align-filter-transfer process. Extensive experiments on three RRSIS benchmarks demonstrate that CMPF achieves state-of-the-art performance. Code is available at: \url{https://github.com/GEK-MM/CMPF}.
\end{abstract}


%
\IEEEpeerreviewmaketitle

\section{Introduction}
%
%
%
%

\begin{figure}[!t]
\centering

\includegraphics[width=\columnwidth, trim=0.5cm 0.7cm 0.6cm 0.6cm, clip]{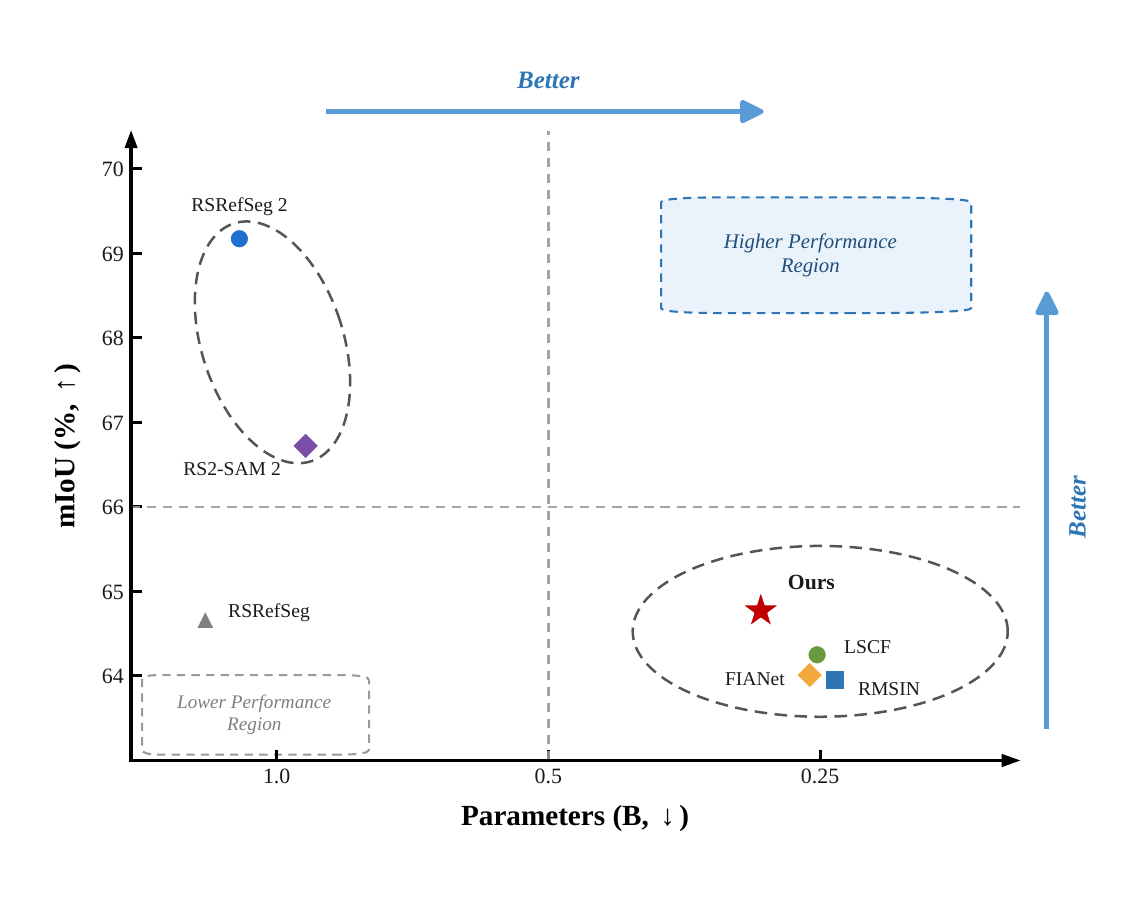}
\caption{Efficiency–accuracy comparison on RRSIS-D test. The parameter axis is logarithmic and reversed, so that fewer parameters appear to the right; arrows indicate better. With 0.29B parameters, CMPF (red star) achieves 64.77 mIoU, the best among lightweight methods, while being roughly one quarter the size of foundation-model-based methods.}
\label{fig:efs}
\end{figure}

\IEEEPARstart{R}{apid} advances in deep learning have driven a wide spectrum of remote sensing (RS) interpretation tasks, evolving from traditional object detection~\cite{cheng2016survey} and change detection~\cite{9955391} to cross-modal vision--language paradigms like visual question answering~\cite{9088993} and visual grounding (RSVG)~\cite{10.1145/3503161.3548316}. However, these cross-modal methods operate at the image or region level, and lack the dense, pixel-level understanding required for fine-grained geospatial analysis. This limitation has driven the emergence of Referring Remote Sensing Image Segmentation (RRSIS)~\cite{10458079}, which transforms static RS imagery into an interactive, high-precision semantic space by demanding language-conditioned, pixel-wise predictions~\cite{11036256}.

Despite its potential for high-stakes applications such as urban planning and emergency response~\cite{11370143}, RRSIS faces unique challenges due to the inherent complexity of RS scenes. Unlike classical RS segmentation, which captures predefined classes within a fixed spatial context, RRSIS requires target localization to be dynamically conditioned on referring expressions. Nevertheless, heavy background clutter~\cite{zheng2020farseg}, extreme multi-scale variations~\cite{niu2021hamanet}, and diverse object orientations~\cite{liu2024rotated, 10458079} make RRSIS particularly challenging, as accurate prediction requires jointly reasoning about categorical semantics and spatial constraints.
 Early RES methods\cite{2016REFSEG_Start,8578700,li2017tracking} employed dual-stream CNN and RNN branches and fused the two modalities by simple concatenation, which enabled basic cross-modal fusion but limited the modeling of long-range dependencies in high-resolution aerial imagery. Transformer-based approaches~\cite{ding2021vlt,kim2022restr,2022lavt} improved global context modeling through self-attention and strengthened vision-language interaction through cross-attention. Recently, foundation-model-based methods~\cite{11243338,11313649} have begun to introduce large-scale pre-trained models into RRSIS, using referring expressions as prompts fed into SAM~\cite{2023SAM} for fine-grained segmentation; these approaches leverage the general capabilities of such models to compensate for the shortcomings of task-specific architectures, attaining high accuracy while incurring the cost of large parameter sizes and high computational overhead. However, most existing multi-scale frameworks propagate or aggregate the cross-modally aligned features across scales without further constraint. Such transfer is indiscriminate, because it preserves both the responses that the expression describes and the activations that it does not. Distracting regions therefore accumulate as the fusion proceeds from coarse to fine. Fig.~\ref{fig:efs}  compares the accuracy of RRSIS-D against model size; with only 0.29B parameters, CMPF achieves the highest mIoU among lightweight methods (64.77) while remaining competitive with models three to four times larger. Although foundation-model-based methods score higher, they require substantially more parameters; consequently, the lightweight regime remains far from saturated.
 
Our approach is centered on preventing the indiscriminate propagation of cross-modal features. Instead of blindly transferring all aligned features, we explicitly filter referring-aware responses prior to cross-scale transmission and selectively fuse them in the decoder, thereby enabling progressively refined segmentation. To handle this issue, we propose the Cross-modal Progressive Filtering (CMPF) network, which converts the propagation of referring information across visual scales from an indiscriminate transfer into a selective one. At every scale, a cross-modal alignment module aligns the visual feature with the language feature of the expression to produce a referring-aware representation; the Referring-aware Filtering (RF) module then suppresses activations not described by the expression, whereas the Spatial Feature Injection (SFI) module injects the filtered higher-level representation into the next scale. In the decoder, the Referring-aware Selective Fusion (RSF) module adaptively fuses multi-scale features to progressively refine the segmentation. In this manner, CMPF prevents irrelevant responses from being propagated across scales, preserves informative referring cues, and thereby improves target localization in complex remote sensing scenes.

Experiments on three benchmarks demonstrate that CMPF achieves state-of-the-art results on RefSegRS \cite{10458079}, RISBench \cite{11036256} and RRSIS-D \cite{liu2024rotated}. 
The main contributions of this study are summarized as follows:
\begin{enumerate}[label=\arabic*), leftmargin=*, widest=3, labelsep=0.5em, align=left]
\item We identify a limitation of existing multi-scale referring image segmentation methods, which propagate cross-modally aligned features across scales without explicit selection. Consequently, both referring-relevant and irrelevant responses are transferred together, allowing distracting information to accumulate during multi-scale fusion.

\item We propose a Cross-modal Progressive Filtering (CMPF) network that propagates referring information selectively across visual scales; CMPF aligns the visual feature with the expression, employs Referring-aware Filtering (RF) to suppress expression-irrelevant activations, and leverages Spatial Feature Injection (SFI) to pass only the informative responses to the next scale. In the decoder stage, a Referring-aware Selective Fusion (RSF) module further fuses the multi-scale features.

\item Experiments on three RRSIS benchmarks validate CMPF, demonstrating superior or highly competitive performance, with the clearest gains at strict IoU thresholds. Ablations and progressive response visualizations confirm that filtering and selective propagation suppress distracting activations, leading to progressively improved localization.

\end{enumerate}

\begin{figure*}
\centering
\includegraphics[width=\textwidth,trim={0cm 0.65cm 0cm 0.65cm}, clip]{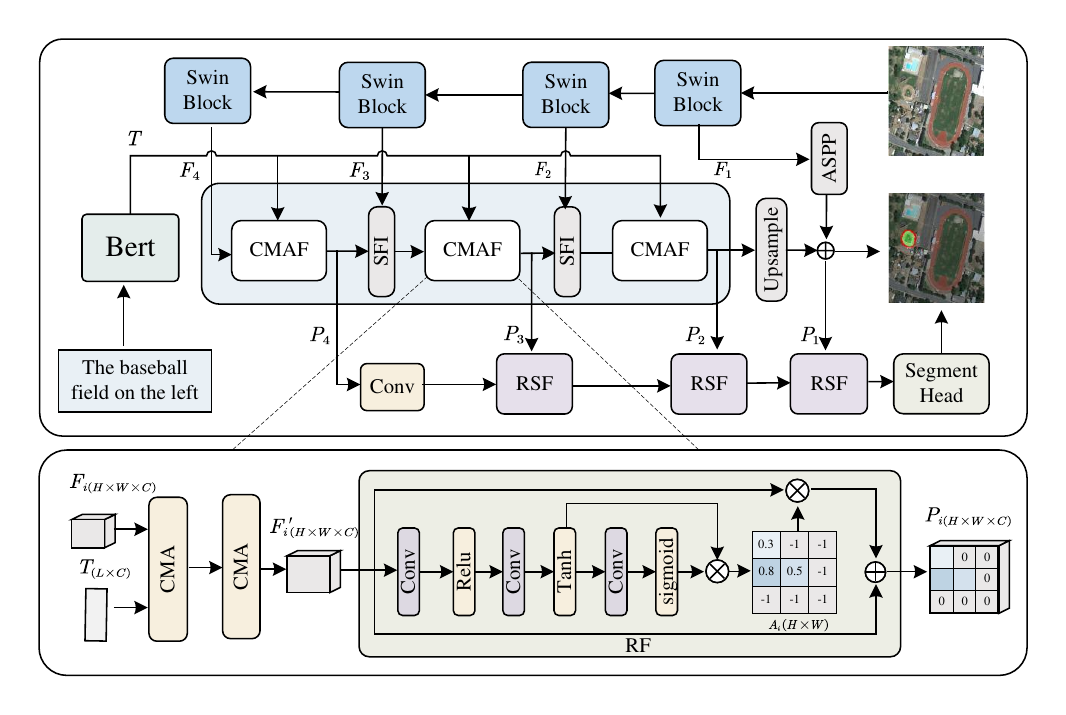}
\caption{Overview of the CMPF architecture for remote sensing referring segmentation. BERT~\cite{devlin-etal-2019-bert} and Swin Transformer~\cite{liu2021swin} extract textual and multi-scale visual features. In the encoder, Cross-modal Alignment (CMA) establishes visual--language correspondence, while Referring-aware Filtering (RF) and Selective Feature Injection (SFI) filter and propagate informative referring features across scales. The decoder employs Referring-aware Selective Fusion (RSF) to aggregate multi-scale features for final mask prediction.}
\label{fig:overview}
\end{figure*}

\section{Related Work}

\subsection{Referring Image Segmentation}

Referring Image Segmentation (RIS) originates in the natural image domain, aiming to segment a target object specified by natural language referring expression.
Early methods primarily employed CNN-RNN architectures with simple concatenative fusion~\cite{2016REFSEG_Start}, while subsequent works introduced tracking-based language specification~\cite{li2017tracking} and dynamic multimodal instance segmentation~\cite{margffoy2018dynamic}.
To strengthen multi-modal interaction, later efforts focused on hierarchical progressive strategies~\cite{9156414, 9430750}, bidirectional relationship inferring~\cite{9157752}, and linguistic structure-guided context modeling~\cite{10.1007/978-3-030-58607-2_4}.

With the advent of Transformers, the field has shifted toward unified feature modeling.
Following the success of the Vision Transformer (ViT)~\cite{dosovitskiy2021vit}, RIS models have increasingly embraced attention-based backbones.
ReSTR~\cite{kim2022restr} leverage vision-language encoders for global context, while LAVT~\cite{2022lavt} adopts early language-aware visual encoding to achieve dense alignment.
This trend is further bolstered by large-scale pre-training via natural language supervision~\cite{2021_CLIP, 9878961}, providing robust cross-modal priors.
To further refine vision--text alignment, PromptRIS~\cite{promptRIS} introduces a prompt-driven paradigm with instance-level contrasting.
To address reasoning complexity, LTS~\cite{jing2021lts} advocates a ``locate-then-segment'' paradigm, and DMMI~\cite{Hu_2023_ICCV} rethinks the standard one-to-one matching.
Recent efforts have explored parameter-efficient tuning~\cite{10377370, DBLP:conf/aaai/HuangXLLHY025}, ambiguity-aware refinement~\cite{mao2025mamba}, and adaptive selection mechanisms~\cite{yue2024ASDA}.
Generalized scenarios (GRES)~\cite{liu2023gres} and foundation model integration~\cite{2023SAM} have also expanded the boundaries of RIS.

\subsection{Remote Sensing Referring Image Segmentation}
RRSIS outputs a pixel-level mask of the region described by a natural-language expression in an aerial image, which is harder than natural-image RIS because of the overhead view, the background clutter, and the extreme scale and orientation variation of ground objects. Yuan~\cite{10458079} pioneered the task with the RefSegRS benchmark and LGCE, a language-guided cross-scale enhancement module for small and scattered objects. RMSIN~\cite{liu2024rotated} released the larger RRSIS-D set and handled rotation and scale with intra- and cross-scale interaction plus adaptive rotated convolution. Recent work is usually evaluated on these two sets and on RISBench.

A second line improves the alignment. FIANet~\cite{10816052} decomposes the expression into object and position fragments before aligning each of them, and CroBIM~\cite{dong2024cross} aligns the two modalities in both directions. LSCF~\cite{11029253} carries language guidance into the decoder to avoid semantic forgetting, while specialized fusion designs~\cite{li2025multimodal,10838614} refine where and how the two modalities are merged. To further enhance semantic consistency, S2CLNet~\cite{11655938} introduces structure-constrained semantic contrastive learning, and CROSS~\cite{10.1007/978-3-032-37435-6_1} leverages cascaded distillation and spatial contrastive learning to address architectural weak-coupling and object-centric semantic bias. Additionally, the Intra and Cross-Scale Feature Interaction Network~\cite{11569979} and the hierarchical vision-language feature fusion framework~\cite{11404179} specifically tackle multi-scale spatial context and complex feature fusion.

A third line changes the backbone or leverages foundation models. Mamba-based interaction has been shown to cut the cost of long-range modeling in natural-image RIS~\cite{wang2025remamba,mao2025mamba}. Recently, Mamba has been introduced to RRSIS, such as the mutual modulation Mamba network (M3RIS)~\cite{11683270}. Furthermore, foundation models are heavily adapted to RRSIS. Some works convert CLIP alignment into SAM prompts~\cite{11243338,11313649} or use SAM-enhanced dual-stream networks~\cite{11666966}. Others transfer SAM2~\cite{rong2026rs2} for customized segmentation. Aside from training-based paradigms, DGL-RSIS~\cite{LI2026105113} proposes a training-free unified framework that decouples global spatial context and local class semantics, demonstrating the potential of transferring vision-language foundation models to the remote sensing domain without additional training.Regardless of the architecture, existing methods align and propagate features across scales without explicit filtering. This allows expression-irrelevant responses to accumulate and contaminate the decoder, hindering precise segmentation. To solve this, CMPF filters the cross-scale messages before transmission and re-estimates the filter at each scale.

\section{Method}
\label{sec:method}

As shown in Fig.~\ref{fig:overview}, the CMPF network encodes the expression into a linguistic feature \(T\) via a pre-trained BERT~\cite{devlin-etal-2019-bert} and extracts multi-scale visual features \(\{F_i\}_{i=1}^{4}\) (from finest \(F_1\) to coarsest \(F_4\)) using a hierarchical Swin Transformer~\cite{liu2021swin}. Within the encoder, each Cross-modal Alignment Filter (CMAF) block utilizes two sequential Cross-Modal Alignment (CMA) modules to align \(T\) with the features at the three coarser scales (\(F_4\) to \(F_2\)). At each transition between scales, the Referring-aware Filtering (RF) module suppresses non-descriptive responses, while the Selective Feature Injection (SFI) module passes only the informative ones to the next finer scale for subsequent alignment and filtering. The finest feature \(F_1\) is processed by an ASPP branch and added to the upsampled output of the last scale to form \(P_1\). Subsequently, the decoder fuses \(\{P_i\}_{i=1}^{4}\) progressively via Referring-aware Selective Fusion (RSF) blocks, followed by a segmentation head that predicts the final target mask.

\subsection{Encoder}

Given the referring-expression feature $T$ and the multi-scale visual features $\{F_i\}$, the encoder performs cross-modal encoding progressively from the coarsest scale to the finest. CMAF is composed of two CMA blocks followed by an RF block. At each stage, CMA applies bidirectional cross-attention between the visual feature and $T$, yielding an aligned multimodal representation. RF then filters the visual output of CMA, and SFI injects the filtered result into the next finer scale. The first two stages stack two CMA blocks, whereas the last stage uses one, so that the encoder yields $\{P_4,P_3,P_2\}$ for the decoder.

\subsubsection{CMA}

The Cross-modal Alignment (CMA) module performs bidirectional feature interaction to align textual semantics with visual representations. Given a visual feature map $\mathbf{F}\in\mathbb{R}^{C\times H\times W}$ and text embeddings $\mathbf{T}\in\mathbb{R}^{N_t\times D}$, we first apply an ASPP layer~\cite{chen2017rethinking} to capture multi-scale visual context, flatten the result, and project it linearly into visual tokens:
\begin{equation}
\mathbf{V}=\mathrm{Proj}_{v}\left(\mathrm{Flatten}\left(\mathrm{ASPP}(\mathbf{F})\right)\right),
\end{equation}
where $\mathbf{V}\in\mathbb{R}^{HW\times D}$ denotes the visual token sequence.

The text embeddings then serve as queries that retrieve semantically relevant visual information: the text tokens attend to the visual tokens through multi-head cross-attention,
\begin{equation}
\widetilde{\mathbf{T}}=\mathbf{T}+\mathrm{MHA}(\mathbf{T},\mathbf{V},\mathbf{V}),
\end{equation}
where $\mathrm{MHA}(\cdot)$ takes the query, key, and value as inputs. The updated textual features are further refined by a GLU-based feed-forward network~\cite{shazeer2020glu}:
\begin{equation}
\mathbf{T}_{\mathrm{out}}=\widetilde{\mathbf{T}}+\mathrm{GLU}(\widetilde{\mathbf{T}}).
\end{equation}

Subsequently, the semantically enhanced text embeddings guide the visual tokens through a reverse cross-attention:
\begin{equation}
\widetilde{\mathbf{V}}=\mathbf{V}+\mathrm{MHA}(\mathbf{V},\mathbf{T}_{\mathrm{out}},\mathbf{T}_{\mathrm{out}}).
\end{equation}
This bidirectional interaction lets the visual features aggregate text-aware semantic cues selectively while keeping their original spatial layout.

Long-range dependencies are further modeled by self-attention, which is applied only at the coarsest stage (stage~4):
\begin{equation}
\mathbf{V}_{s}=\widetilde{\mathbf{V}}+\mathrm{MHA}(\widetilde{\mathbf{V}},\widetilde{\mathbf{V}},\widetilde{\mathbf{V}}).
\end{equation}
At the higher-resolution stages~2 and~3 the operation is omitted, because dense spatial tokens would make its cost quadratic; hence $\mathbf{V}_{s}=\widetilde{\mathbf{V}}$ there.

Finally, the aligned visual tokens pass through another GLU-based feed-forward network and are projected back to the spatial domain:
\begin{equation}
\mathbf{F}_{\mathrm{out}}=\mathrm{Reshape}\left(\mathrm{Proj}_{o}\left(\mathbf{V}_{s}+\mathrm{GLU}(\mathbf{V}_{s})\right)\right),
\end{equation}
where $\mathbf{F}_{\mathrm{out}}\in\mathbb{R}^{C\times H\times W}$ has the same spatial resolution as the input feature map. Bidirectional cross-modal interaction combined with stage-selective self-attention thus gives CMA effective semantic alignment at a manageable computational cost.

\subsubsection{Referring-aware Filtering (RF)}

To suppress responses from irrelevant regions while enhancing those associated with the referring expression, we devise a Referring-aware Filtering (RF) module. Given an input feature map $\mathbf{V}_i \in \mathbb{R}^{C \times H \times W}$, RF first transforms it into an intermediate spatial response through a lightweight convolutional mapping:
\begin{equation}
\mathbf{Z}_i =
\mathrm{Conv}_2\left(
\mathrm{ReLU}\left(
\mathrm{Conv}_1(\mathbf{V}_i)
\right)
\right).
\end{equation}

The intermediate feature $\mathbf{Z}_i$ is subsequently activated by $\mathrm{Tanh}(\cdot)$. The resulting response is retained in one branch and fed into a convolution followed by a sigmoid activation in the other branch. Their element-wise product yields the final spatial filter:
\begin{equation}
\mathbf{A}_i =
\mathrm{Tanh}(\mathbf{Z}_i)
\odot
\sigma\left(
\mathrm{Conv}_3\left(
\mathrm{Tanh}(\mathbf{Z}_i)
\right)
\right),
\label{eq:modulation}
\end{equation}
where $\mathbf{A}_i \in \mathbb{R}^{H \times W}$, $\sigma(\cdot)$ denotes the sigmoid function, and $\odot$ denotes element-wise multiplication. The $\mathrm{Tanh}$ branch retains signed responses, whereas the sigmoid branch adaptively modulates their contribution.

Finally, the learned filter is applied to the input feature map through a residual formulation:
\begin{equation}
\mathbf{P}_i =
\mathbf{V}_i +
\mathbf{V}_i \odot \mathbf{A}_i,
\label{eq:residual}
\end{equation}
where $\mathbf{P}_i \in \mathbb{R}^{C \times H \times W}$ denotes the refined output feature. During multiplication, $\mathbf{A}_i$ is broadcast along the channel dimension. This residual filtering mechanism selectively enhances expression-relevant regions while preserving the original visual representation.

\begin{figure}[!t]
\centering
\includegraphics[width=\columnwidth, trim=0cm 0.7cm 0.6cm 0.6cm, clip]{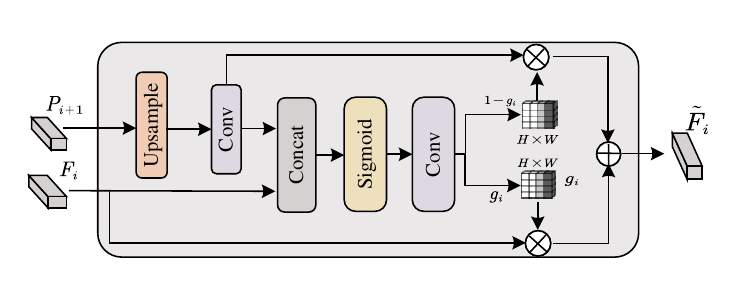}
\caption{Architecture of the proposed Spatial Feature Injection (SFI) module. The complementary gate $g_i$ adaptively balances, at each spatial location, the current-scale detail $\mathbf{F}_i$ and the upsampled high-level semantic feature $\mathbf{P}_{i+1}$.}
\label{fig:sfi}
\end{figure}

\begin{figure}[!t]
\centering
\includegraphics[width=\columnwidth, trim=0.3cm 0.6cm 0.2cm 0.4cm, clip]{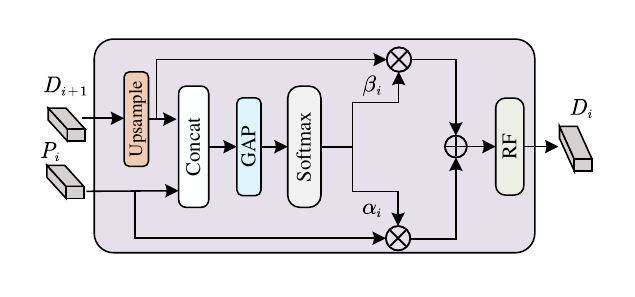}
\caption{Architecture of the RSF module. A Softmax-normalized pair of competitive weights $(\alpha_i,\beta_i)$ is learned to adaptively balance the spatial details in $\mathbf{P}_i$ against the semantic context of the upsampled $\mathbf{D}_{i+1}$, after which the fused representation is refined.}
\label{fig:rsf}
\end{figure}

\subsubsection{Spatial Feature Injection (SFI)}

SFI is illustrated in Fig.~\ref{fig:sfi}. To progressively propagate high-level semantic cues to finer-resolution visual representations, we devise a Spatial Feature Injection (SFI) module. Given the current-stage feature $\mathbf{F}_i \in \mathbb{R}^{C_i \times H_i \times W_i}$ and the output $\mathbf{P}_{i+1}$ from the preceding coarser stage, SFI first upsamples $\mathbf{P}_{i+1}$ and projects it to the channel dimension of $\mathbf{F}_i$:
\begin{equation}
\widehat{\mathbf{P}}_i =
\mathrm{Conv}_{p}\left(
\mathrm{Up}(\mathbf{P}_{i+1})
\right),
\end{equation}
where $\widehat{\mathbf{P}}_i \in \mathbb{R}^{C_i \times H_i \times W_i}$.

Rather than directly adding features from different stages, SFI estimates a content-adaptive complementary gate from their concatenation:
\begin{equation}
\mathbf{g}_i =
\sigma\left(
\mathrm{Conv}_{2}\left(
\mathrm{ReLU}\left(
\mathrm{Conv}_{1}\left(
\operatorname{Concat}
(\mathbf{F}_i,\widehat{\mathbf{P}}_i)
\right)
\right)
\right)
\right),
\end{equation}
where $\operatorname{Concat}(\cdot,\cdot)$ denotes channel-wise concatenation and $\mathbf{g}_i \in [0,1]^{C_i \times H_i \times W_i}$.

The gate and its complement are then used to balance fine-grained visual details against high-level semantic cues:
\begin{equation}
\widetilde{\mathbf{F}}_i =
\mathbf{g}_i \odot \mathbf{F}_i
+
(1-\mathbf{g}_i) \odot \widehat{\mathbf{P}}_i,
\end{equation}
where $\odot$ denotes element-wise multiplication. A large value of $\mathbf{g}_i$ assigns greater importance to the local details in $\mathbf{F}_i$, whereas a small value promotes the injection of high-level semantics from $\widehat{\mathbf{P}}_i$. In this way, the complementary gating mechanism adaptively reconciles semantic guidance and spatial detail across scales without introducing redundant feature responses.

\subsection{Decoder}

\subsubsection{RSF Module}

The architecture of the RSF module is presented in Fig.~\ref{fig:rsf}. To progressively recover spatial details while preserving high-level semantics, we employ an RSF module to fuse features from adjacent decoding stages. At the $i$-th stage, $\mathbf{P}_i$ denotes the projected feature at the current resolution, and $\mathbf{D}_{i+1}$ denotes the decoded feature from the preceding coarser stage. We first upsample $\mathbf{D}_{i+1}$ to match the spatial resolution of $\mathbf{P}_i$:
\begin{equation}
\overline{\mathbf{D}}_i =
\mathrm{Up}(\mathbf{D}_{i+1}).
\end{equation}

Rather than directly summing the two features, RSF derives a pair of competitive fusion weights from their global context:
\begin{equation}
[\alpha_i,\beta_i] =
\mathrm{Softmax}\left(
\mathrm{Conv}_{1\times1}\left(
\mathrm{GAP}\left(
\operatorname{Concat}(\mathbf{P}_i,\overline{\mathbf{D}}_i)
\right)
\right)
\right),
\end{equation}
where $\mathrm{GAP}(\cdot)$ denotes global average pooling. The Softmax operation enforces $\alpha_i+\beta_i=1$, so that current-scale details and high-level semantic information compete for their relative contributions rather than being independently amplified.

The fused representation is subsequently refined as:
\begin{equation}
\mathbf{D}_i =
RF\left(
\alpha_i \odot \mathbf{P}_i +
\beta_i \odot \overline{\mathbf{D}}_i
\right),
\end{equation}
Consequently, RSF adaptively balances fine-grained appearance details and coarse semantic context at each decoding stage.

After three successive RSF operations, the finest decoded feature is transformed into a two-channel segmentation logit map by a lightweight prediction head:
\begin{equation}
\mathbf{M}=
\mathrm{Conv}_{1\times1}\left(
\mathrm{GELU}\left(
\mathrm{GN}\left(
\mathrm{Conv}_{3\times3}(\mathbf{D}_0)
\right)
\right)
\right).
\end{equation}
The logits are bilinearly upsampled to the input resolution, and the final mask is obtained by selecting the class with the maximum prediction score at each pixel.

\begin{table*}[t]
\centering
\caption{EXPERIMENTAL RESULTS ON THE \textbf{RRSIS-D\cite{liu2024rotated}} TESTING DATASET. THE SYMBOL \(\dagger\) INDICATES THE RESULTS ARE QUOTED FROM \cite{11029253}, THE SYMBOL \(*\) INDICATES THAT THE RESULTS ARE QUOTED FROM \cite{dong2024cross}.  THE SYMBOL \(\ddagger\) INDICATES THAT THE RESULTS ARE QUOTED FROM \cite{11404179}.}
\label{tab:rrsis_d_final}
\small
\renewcommand{\arraystretch}{1} 
\setlength{\tabcolsep}{8pt} 
\begin{tabular}{@{}l c c c c c c c c c c@{}}
\toprule
Methods & Publication & Vision & Language & Pr@50 & Pr@60 & Pr@70 & Pr@80 & Pr@90 & oIoU & mIoU \\ 
\midrule
LAVT \cite{2022lavt}\(^{\dagger}\) & CVPR 2022 & Swin-B & BERT & 69.67 & 63.87 & 54.14 & 42.02 & 24.96 & 77.66 & 61.25 \\
DMMI \cite{Hu_2023_ICCV}\(^{*}\) & ICCV 2023 & Swin-B & BERT & 68.74 & 60.96 & 50.33 & 38.38 & 21.63 & 76.20 & 60.12 \\
RefSegformer \cite{10460426}\(^{*}\) & TIP 2024 & Swin-B & BERT & 66.59 & 59.58 & 49.93 & 33.78 & 23.30 & 77.40 & 58.99 \\
CrossVLT \cite{10345690}\(^{*}\) & T-MM 2024 & Swin-B & BERT & 66.42 & 59.41 & 49.76 & 38.67 & 23.30 & 75.48 & 58.48 \\
LGCE \cite{10458079}\(^{\dagger}\) & TGRS 2024 & Swin-B & BERT & 69.90 & 63.96 & 53.56 & 40.81 & 24.21 & 77.23 & 60.91 \\
RMSIN \cite{liu2024rotated}\(^{\dagger}\) & CVPR 2024 & Swin-B & BERT & 74.15 & 67.08 & 55.64 & 41.65 & 23.07 & 77.49 & 63.95 \\
FIANet \cite{10816052}\(^{\dagger}\) & TGRS 2025 & Swin-B & BERT & 74.46 & 66.96 & 56.31 & 42.83 & 24.13 & 76.91 & 64.01 \\
LSCF \cite{11029253} & TGRS 2025 & Swin-B & BERT & 74.30 & \underline{67.69} & \underline{56.32} & 43.08 & \underline{25.67} & 77.42 & 64.25 \\
VSPNet \cite{11404179}\(^{\ddagger}\) & TGRS 2026 & VMamba & CLIP & \textbf{75.06} & 67.58 & 55.42 & \underline{44.61} & 25.43 & \textbf{77.81} & \underline{64.69} \\

\midrule
\textbf{CMPF} & Ours & Swin-B & BERT & \underline{74.60} & \textbf{68.28} & \textbf{57.80} & \textbf{44.96} & \textbf{26.54} & \underline{77.79} & \textbf{64.77} \\
\bottomrule
\end{tabular}
\end{table*}

\section{Experiments}

\subsection{Datasets and Evaluation Metrics}

\subsubsection{Datasets}
To evaluate the performance of our proposed method for the Referring Remote Sensing Image Segmentation (RRSIS) task, we conduct our experiments on three benchmark datasets: RefSegRS \cite{10458079}, RRSIS-D \cite{liu2024rotated}, and RISBench \cite{11036256}.

\textbf{RefSegRS}\cite{10458079} is the first dataset dedicated to RRSIS, comprises 4,420 image-text-mask triplets, with 2,172 assigned to training, 431 to validation, and the remaining 1,817 to testing. Images are resized to $512 \times 512$ at 0.13\,m resolution, covering 14 categories (e.g., roads, vehicles, buildings) with template-based referring expressions.

\textbf{RRSIS-D}\cite{liu2024rotated} is a larger benchmark containing 17,402 image-text pairs at $800 \times 800$ resolution, split into 12,181 for training, 1,740 for validation, and 3,481 for testing. It includes 20 semantic categories and 7 attribute types, with masks initially produced by SAM and then manually refined.

\textbf{RISBench}\cite{11036256} provides 52,472 triplets at $512 \times 512$ with resolutions from 0.1\,m to 30\,m. covers 26 categories, each with 8 attributes, split into 26,300 for training, 10,013 for validation, and 16,149 for testing, enabling broad evaluation of RRSIS generalization.

\subsubsection{Evaluation Metrics and Training Details}
Following established research in the RRSIS field, we employ overall Intersection-over-Union (oIoU), mean Intersection-over-Union (mIoU), and Precision at various thresholds (Pr@$X$) as the primary evaluation metrics.

\begin{table*}[t]
\centering
\caption{EXPERIMENTAL RESULTS ON THE \textbf{REFSEGRS\cite{10458079}} TESTING DATASET. THE SYMBOL \(\dagger\) INDICATES THE RESULTS ARE QUOTED FROM \cite{11029253}, THE SYMBOL \(*\) INDICATES THE RESULTS ARE QUOTED FROM \cite{dong2024cross}. THE SYMBOL \(\ddagger\) INDICATES THE RESULTS ARE QUOTED FROM \cite{11404179}.}
\label{tab:refsegrs_final}
\small
\renewcommand{\arraystretch}{1} 
\setlength{\tabcolsep}{8pt} 
\begin{tabular}{@{}l c c c c c c c c c c@{}} 
\toprule
Methods & Publication & Vision & Language & Pr@50 & Pr@60 & Pr@70 & Pr@80 & Pr@90 & oIoU & mIoU \\ 
\midrule

LAVT \cite{2022lavt}\(^{\dagger}\) & CVPR 2022 & Swin-B & BERT & 63.37 & 48.73 & 27.17 & 12.71 & 3.69 & 75.04 & 52.89 \\
DMMI \cite{Hu_2023_ICCV}\(^{*}\) & ICCV 2023 & Swin-B & BERT & 63.89 & 44.30 & 19.81 & 6.49 & 1.00 & 68.58 & 52.15 \\
RefSegformer \cite{10460426}\(^{*}\) & TIP 2024 & Swin-B & BERT & 50.25 & 28.01 & 17.83 & 9.19 & 2.48 & 71.13 & 47.12 \\
CrossVLT \cite{10345690}\(^{*}\) & T-MM 2024 & Swin-B & BERT & 41.94 & 25.43 & 15.19 & 3.71 & 1.76 & 69.73 & 42.81 \\
LGCE \cite{10458079}\(^{\dagger}\) & TGRS 2024 & Swin-B & BERT & 74.04 & 61.39 & 39.71 & 17.71 & 5.17 & 77.71 & 59.59 \\
RMSIN \cite{liu2024rotated}\(^{\dagger}\) & CVPR 2024 & Swin-B & BERT & 71.07 & 57.26 & 34.87 & 13.15 & 2.75 & 72.68 & 57.91 \\
FIANet \cite{10816052}\(^{\dagger}\) & TGRS 2025 & Swin-B & BERT & 83.60 & 75.18 & 59.55 & 30.65 & 6.16 & 77.74 & 67.86 \\
LSCF \cite{11029253} & TGRS 2025 & Swin-B & BERT & \underline{87.51} & \underline{82.89} & \underline{75.85} & \underline{62.65} & \underline{33.77} & \underline{83.27} & \underline{77.44} \\ 
VSPNet \cite{11404179}\(^{\ddagger}\)  & TGRS 2026 & VMamba & CLIP & 85.12 & 78.59 & 63.01 & 35.42 & 9.23 & 79.60 & 69.59 \\

\midrule
\textbf{CMPF} & Ours & Swin-B & BERT & \textbf{88.33} & \textbf{83.82} & \textbf{76.72} & \textbf{64.78} & \textbf{44.36} & \textbf{84.03} & \textbf{79.01} \\
\bottomrule
\end{tabular}
\end{table*}

\begin{figure*}[!t]
\centering
\includegraphics[width=\textwidth, trim=0.5cm 0.5cm 0.5cm 0cm, clip]{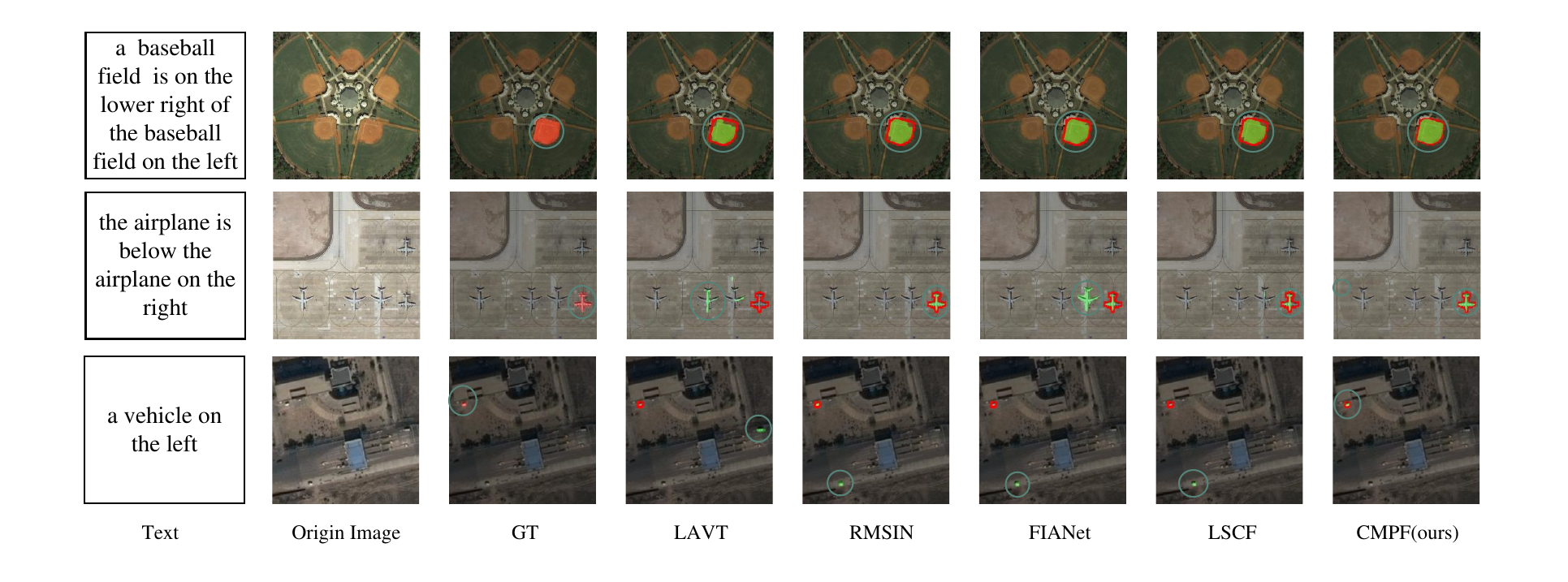}
\caption{Comparison of visualization results. referring expressions are listed on the left, followed by the original images, ground truth (GT), and results from the compared methods. The results for FIANet are obtained from the official model files, while those for the others are generated through our re-implemented version. {\color{red}Red }boundaries denote the GT. {\color{green}green }boundaries denote segmentation results.}
\label{fig:comparison}
\end{figure*}

\subsection{SOTA Comparisons}

Following most of the SOTA approaches \cite{liu2024rotated,11029253,10816052}, we implement our model using PyTorch and train it end-to-end with the AdamW optimizer. The visual backbone employs Swin-Base \cite{liu2021swin} initialized with pre-trained weights from ImageNet-22K, while text features are extracted using BERT-Base-Uncased \cite{devlin-etal-2019-bert}. The learning rate is set to $3\times10^{-5}$ for RRSIS-D \cite{liu2024rotated} and RISBench \cite{11036256}, and $5\times10^{-5}$ for RefSegRS \cite{10458079} to facilitate faster adaptation. All input images are resized to $512\times512$. No data augmentation techniques are applied during training. All experiments are conducted on $2 \times$ RTX 3090 GPUs. The batch size is set to 8 per GPU for RRSIS-D \cite{liu2024rotated} and RISBench \cite{11036256}, and 4 per GPU for RefSegRS \cite{10458079}.

The proposed CMPF is supervised with a compound loss function:
\begin{equation}
\mathcal{L} = 0.5 \cdot \mathcal{L}_{\text{CE}} + 0.5 \cdot \mathcal{L}_{\text{Dice}},
\end{equation}
where $\mathcal{L}_{\text{CE}}$ denotes cross-entropy loss and $\mathcal{L}_{\text{Dice}}$ denotes Dice loss. Both loss components are computed directly on the final predicted mask across all three benchmark datasets.

\begin{table*}[t]
\centering
\caption{COMPARATIVE EXPERIMENTAL RESULTS ON THE \textbf{RISBENCH\cite{11036256}} DATASET. THE SYMBOL \(\dagger\) INDICATES THAT THE RESULTS ARE QUOTED FROM \cite{11029253}, AND THE SYMBOL \(*\) INDICATES THAT THE RESULTS ARE QUOTED FROM \cite{dong2024cross}.}
\label{tab:results_final}
\small
\renewcommand{\arraystretch}{1} 
\setlength{\tabcolsep}{8pt} 
\begin{tabular}{@{}l c c c c c c c c c c@{}} 
\toprule
Methods & Publication & Vision & Language & Pr@50 & Pr@60 & Pr@70 & Pr@80 & Pr@90 & oIoU & mIoU \\ 
\midrule
LAVT \cite{2022lavt}\(^{\dagger}\) & CVPR 2022 & Swin-B & BERT & 71.33 & 66.32 & 59.37 & 48.64 & 28.84 & 74.71 & 63.78 \\
DMMI \cite{Hu_2023_ICCV}\(^{*}\) & ICCV 2023 & Swin-B & BERT & 72.05 & 66.48 & 59.07 & 47.16 & 26.57 & 74.82 & 63.93 \\
RefSegformer \cite{10460426}\(^{*}\) & TIP 2024 & Swin-B & BERT & 69.15 & 63.24 & 55.33 & 43.27 & 24.20 & 74.23 & 61.25 \\

RMSIN \cite{liu2024rotated}\(^{\dagger}\) & CVPR 2024 & Swin-B & BERT & 74.54 & 69.51 & 62.32 & 51.53 & 30.58 & 73.35 & 66.86 \\
LGCE \cite{10458079}\(^{\dagger}\) & TGRS 2024 & Swin-B & BERT & 71.19 & 65.92 & 58.94 & 47.87 & 27.86 & 74.34 & 63.39 \\
FIANet \cite{10816052}\(^{\dagger}\) & TGRS 2025 & Swin-B & BERT & 74.80 & 69.86 & 63.16 & 52.70 & 31.87 & 74.23 & 66.83 \\
LSCF \cite{11029253} & TGRS 2025 & Swin-B & BERT & \underline{76.08} & \underline{71.29} & \underline{64.96} & \underline{55.13} & \underline{36.73} & \underline{74.88} & \underline{68.53} \\ 
\midrule
\textbf{CMPF} & Ours & Swin-B & BERT & \textbf{77.18} & \textbf{72.69} & \textbf{66.69} & \textbf{57.66} & \textbf{40.34} & \textbf{75.26} & \textbf{70.02} \\
\bottomrule
\end{tabular}
\end{table*}

We compare the proposed CMPF with classical and recent methods in RIS and RRSIS, including  LAVT~\cite{2022lavt}, DMMI~\cite{Hu_2023_ICCV}, CrossVLT~\cite{10345690}, RefSegFormer~\cite{10460426}, LGCE~\cite{10458079}, RMSIN~\cite{liu2024rotated}, FIANet~\cite{10816052}, LSCF~\cite{11029253} and VSPNet~\cite{11404179} .

The experiments are conducted on three benchmarks: RRSIS-D, RefSegRS, and RISBench, which cover early CNN--RNN architectures, CLIP-based transfer learning methods, and recent SOTA Swin-BERT frameworks.
For fair comparisons, the results of existing methods are mainly taken from~\cite{11029253} and~\cite{dong2024cross} under the same evaluation settings.
We report mIoU, oIoU, and Pr@, and present the results on the three datasets separately below.

\textbf{Quantitative Results and Analysis.}
Tables~\ref{tab:results_final},~\ref{tab:rrsis_d_final}, and~\ref{tab:refsegrs_final} compare CMPF with existing methods on RISBench~\cite{11036256}, RRSIS-D~\cite{liu2024rotated}, and RefSegRS~\cite{10458079}, respectively. CMPF achieves the best performance on every reported test metric across all three datasets.

On RISBench, CMPF outperforms LSCF~\cite{11029253} by 1.10\% in Pr@50, 3.61\% in Pr@90, 0.38\% in oIoU, and 1.49\% in mIoU. The more pronounced gain at Pr@90 indicates improved mask quality under a strict overlap criterion.

On RRSIS-D, CMPF achieves the best test-set performance on every metric, reaching 74.60\% Pr@50, 26.54\% Pr@90, 77.79\% oIoU, and 64.77\% mIoU. Compared with the previous best results, CMPF improves Pr@70 and Pr@80 by 1.48\% and 1.88\%, respectively, demonstrating robust target localization at increasingly stringent IoU thresholds.

On RefSegRS, CMPF obtains 88.33\% Pr@50, 44.36\% Pr@90, 84.03\% oIoU, and 79.01\% mIoU on the test set. In particular, it surpasses LSCF by 10.59\% at Pr@90 and 1.57\% in mIoU. These consistent improvements, especially at strict IoU thresholds, validate the effectiveness of selectively filtering and propagating referring-aware features across scales.

\textbf{Comparison of visualization results.} Fig.~\ref{fig:comparison} qualitatively compares CMPF with state-of-the-art methods in four remote sensing scenarios. In dense textures (\textit{e.g.}, ``a ship on the lower right''), baselines suffer from under-segmentation or incomplete boundaries, while CMPF produces masks closely matching the ground truth. Under complex spatial relations and distracting contexts, LAVT is confused by densely arranged airplanes and RMSIN misses targets, whereas CMPF correctly decodes cascaded spatial references. For micro-scale targets in cluttered backgrounds, previous methods yield false alarms due to structural shadows, while CMPF suppresses noise effectively. These results validate that CMPF mitigates spatial collapse and achieves robust, precise segmentation.

\subsection{Ablation Studies}
To verify the effectiveness of each proposed component and our design choices, we conduct extensive ablation experiments on the RRSIS-D\cite{liu2024rotated} and RISBench\cite{11036256} dataset. All ablation models are trained under the same experimental settings to ensure a fair comparison.

\begin{table}[t]
\centering
\caption{\small ABLATION STUDY OF DIFFERENT COMPONENTS ON RRSISD\cite{liu2024rotated} VAL DATASET.}
\label{tab:ablation}
\small 
\renewcommand{\arraystretch}{1.1}
\setlength{\tabcolsep}{0pt} 
\begin{tabularx}{\columnwidth}{@{\extracolsep{\fill}}cccc|ccc|cc@{}}
\toprule
CMA & SFI & RF & RSF & Pr@50 & Pr@70 & Pr@90 & oIoU & mIoU \\ \midrule
\multicolumn{4}{c|}{Baseline}
& 72.84 & 55.10 & 25.01 & 74.52 & 62.81 \\ \midrule
\checkmark & & & & 73.71 & 56.02 & 25.68 & 75.24 & 63.65 \\
\checkmark & & \checkmark & & 74.11 & 57.53 & 26.32 & 75.90 & 64.47 \\
\checkmark & \checkmark & & & 74.94 & 58.74 & 27.07 & 76.87 & 65.61 \\
\checkmark & & & \checkmark & 75.46 & 58.22 & \textbf{27.64} & \textbf{76.97} & 65.33 \\
\checkmark & \checkmark & \checkmark & \checkmark &
\textbf{75.63} & \textbf{59.77} & 27.13 &
76.95 & \textbf{66.06} \\ \bottomrule
\end{tabularx}

\end{table}

\begin{table}[]
\centering
\caption{\normalsize Ablation Study of Inter-scale Visual Feature Fusion Strategies in the Encoder (SFI) and Decoder (RSF) on the RRSIS-D Validation.}
\label{tab:ablation_fusion}
\small
\renewcommand{\arraystretch}{1.3}
\setlength{\tabcolsep}{0pt}
\begin{tabular*}{\columnwidth}{@{\extracolsep{\fill}}l *{4}{c}@{}}
\toprule
 & \multicolumn{2}{c}{Encoder (SFI)} & \multicolumn{2}{c}{Decoder (RSF)} \\
\cmidrule(lr){2-3} \cmidrule(lr){4-5}
Fusion Strategies & oIoU & mIoU & oIoU & mIoU \\ \midrule
Element-wise Sum & 76.75 & 65.50 & 76.95 & 65.15 \\
Element-wise Mul & 76.87 & 65.36 & 76.87 & 65.19 \\
Gated Fusion     & 76.25 & 64.81 & 76.90 & 65.30 \\
Concat+Conv      & \textbf{76.99} & 65.33 & \textbf{77.65} & 65.76 \\
\textbf{(Ours)} & 76.95 & \textbf{66.06} & 76.95 & \textbf{66.06} \\ \bottomrule
\end{tabular*}
\end{table}



\begin{figure*}[!htbp]
\centering
\includegraphics[width=\textwidth]{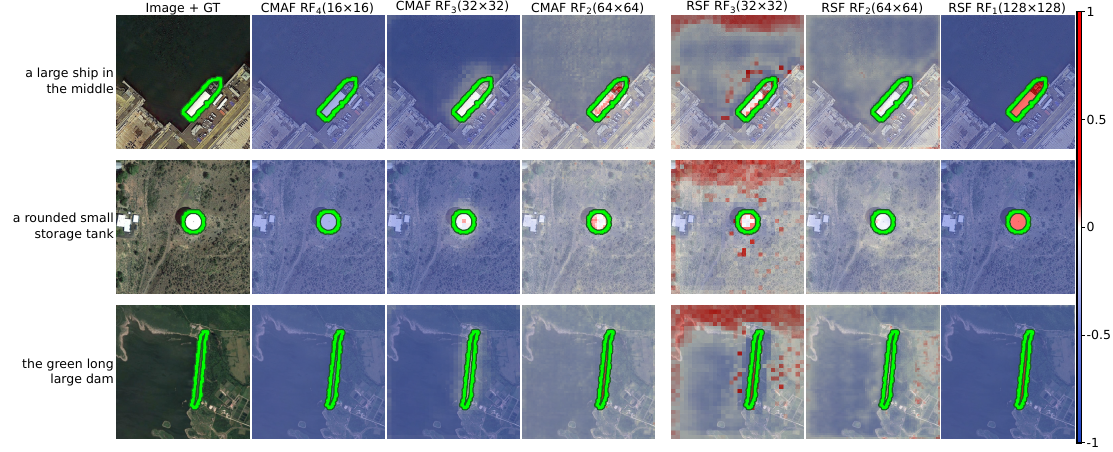}
\caption{Modulation maps $A$ are included in the encoder CMAF modules (RF$_4$--RF$_2$, 16$\times$16--64$\times$64) and the decoder RSF modules (RF$_3$--RF$_1$, 32$\times$32--128$\times$128) for three referring expressions, where red indicates enhancement ($A>0$) and blue indicates suppression ($A<0$), with white being neutral ($A=0$); GT are blended with the input image, and the ground-truth contour is drawn in green.
}
\label{fig:visual block}
\end{figure*}

\textbf{Effectiveness of Components.} Table~\ref{tab:ablation} reports the ablation results on the RRSIS-D validation set. CMA alone increases mIoU from 62.81\% to 63.65\% (+0.84\%), while Pr@50, Pr@70, Pr@90, and oIoU improve by 0.87\%, 0.92\%, 0.67\%, and 0.72\%, respectively, demonstrating the benefit of bidirectional cross-modal alignment. Adding RF further improves mIoU to 64.47\% (+0.82\% over CMA), with the largest gain at Pr@70 (+1.51\%), indicating that suppressing irrelevant responses yields more discriminative representations. However, equipping CMA with SFI produces a larger gain, reaching 65.61\% mIoU (+1.96\% over CMA) and improving Pr@70 by 2.72\%, confirming the effectiveness of cross-scale Spatial Feature Injection. Alternatively, CMA+RSF improves mIoU to 65.33\% (+1.68\% over CMA) and achieves the best Pr@90 (27.64\%) and oIoU (76.97\%) in the table, showing that adaptive multi-scale aggregation benefits mask overlap and high-threshold localization. 

The two modules thus play complementary roles: SFI improves inter-scale feature propagation, while RSF improves decoder fusion. By integrating all components, CMPF achieves the best Pr@50 (75.63\%), Pr@70 (59.77\%), and mIoU (66.06\%), an overall improvement of 3.25\% over the baseline. Relative to the strongest partial model in mIoU (CMA+SFI), the full model gains 0.45\% mIoU and 1.03\% Pr@70, while its Pr@90 and oIoU are only 0.51\% and 0.02\% below the best CMA+RSF values. This slight high-threshold trade-off does not change the conclusion: the full model delivers the best average segmentation quality, while CMA+RSF remains marginally preferable when very strict localization is the sole criterion.

\textbf{Effectiveness of SFI and RSF.} Table~\ref{tab:ablation_fusion} evaluates SFI and RSF on the RRSIS-D validation set by replacing each with four alternative inter-scale fusion strategies. SFI uses a per-pixel/per-channel gate to inject deeper referring-aware features into shallower ones, allowing each position to balance local detail and injected semantics. Replacing SFI with alternatives drops mIoU by 0.56--1.25 points, with the largest drop for global-statistics-based gated fusion, showing that the spatially dense gate accounts for the gain. RSF predicts channel-wise weights from adjacent scales' global context and applies $3\times3$ refinement; replacing it decreases mIoU by 0.30--0.91 points, indicating fixed summation or concatenation mis-allocates scale contributions. For oIoU, SFI and RSF attain 76.95, comparable to element-wise/gated baselines. Although Concat+Conv achieves the highest oIoU (77.65) by enlarging the fused response (covering more target), it leaves per-pixel semantics unchanged, explaining its lower mIoU (65.76 vs. 66.06).

\textbf{Effectiveness across Expression Lengths.} Table~\ref{tab:length_ablation} reports the segmentation performance under different referring expression lengths across RRSIS-D, RefSegRS, and RISBench. Benefiting from the proposed progressive align-filter-transfer mechanism (CMA+SFI), CMPF exhibits strong robustness across varying expression lengths. On RRSIS-D, as the expression length increases from \([0,5]\) to \([6,9]\), mIoU improves from 65.49\% to 72.27\% on the Val set, demonstrating that our model effectively leverages richer semantic cues to refine cross-scale filtering. Notably, although performance drops to 49.72\% for long expressions (\([10,20]\)), the SFI module still suppresses expression-irrelevant responses during cross-scale propagation, preventing the accumulation of background noise. For RISBench, performance steadily improves with length, reaching 71.99\% (Test) for \([21,+)\) words, which proves that our progressive filtering mechanism can accurately localize targets even under complex and verbose descriptions. Furthermore, on RefSegRS, which is dominated by short expressions, CMPF achieves the highest Test mIoU of \textbf{82.21}\%. Overall, by explicitly distinguishing which information should be transferred and filtered at each scale, CMPF successfully overcomes the interference of redundant information in long texts and the semantic insufficiency in short texts.

\begin{table}[]
\centering
\caption{\normalsize Segmentation Performance (mIoU, \%) under Different Referring Expression Lengths on RRSIS-D, RefSegRS and RISBench.}
\label{tab:length_ablation}
\small
\renewcommand{\arraystretch}{1.3}
\setlength{\tabcolsep}{0pt}
\begin{tabularx}{\columnwidth}{@{\extracolsep{\fill}}lcccccc@{}}
\toprule
\multirow{2}{*}{Length (words)} & \multicolumn{2}{c}{RRSIS-D} & \multicolumn{2}{c}{RefSegRS} & \multicolumn{2}{c}{RISBench} \\
\cmidrule(lr){2-3} \cmidrule(lr){4-5} \cmidrule(lr){6-7}
 & Val & Test & Val & Test & Val & Test \\ \midrule
\([0,5]\)    & 65.49 & 63.50 & 91.61 & 78.60 & 60.71 & \textbf{85.37} \\
\([6,9]\)    & 72.27 & \textbf{72.22} & \textbf{94.72} & 82.21 & 67.24 & 67.59 \\
\([10,20]\)  & 49.72 & 48.18 & --    & --    & \textbf{69.78} & 70.15 \\
\([21,+)\)   & --    & --    & --    & --    & 69.52 & 71.99 \\ \midrule
All          & \textbf{66.06} & 64.77 & 91.93 & 79.01 & 69.55 & 70.02 \\
\bottomrule
\end{tabularx}
\end{table}

\subsection{Visualisation of Spatial filter.}   

Fig.~\ref{fig:visual block} visualizes the feature modulation performed by the encoder CMAF modules and the decoder RSF modules via the residual formulation in Eq.~(\ref{eq:residual}), $\mathbf{P}_i = \mathbf{V}_i + \mathbf{V}_i \odot \mathbf{A}_i$, with the modulation map $\mathbf{A}_i$ computed as in Eq.~(\ref{eq:modulation}). Determined solely by the sign of $\mathbf{A}_i$, this modulation effect amplifies the response when $\mathbf{A}_i>0$, suppresses it when $\mathbf{A}_i<0$, and leaves it unchanged when $\mathbf{A}_i=0$; $\mathbf{A}_i$ is thus a signed, spatially resolved coefficient of enhancement or suppression. Accordingly, the encoder CMAF modules (RF$_4$--RF$_2$, 16$\times$16--64$\times$64) and the decoder RSF modules (RF$_3$--RF$_1$, 32$\times$32--128$\times$128) are presented on a common scale $\mathbf{A}_i\in[-1,1]$.

The figure shows that the encoder discards, at each scale, the content irrelevant to the expression, whereas the decoder retains the candidate target regions among the filtered and fused features. The encoder filtering is content-specific: F$_4$ turns the water, the wharf with containers, and the vegetation blue, whereas the hull keeps a weak positive response; only at finer scales does it separate the object: for the small tank and the thin dam, CMAF RF$_4$ is almost entirely blue, because a 16$\times$16 cell spans extended irrelevant areas and leaves the separation to CMAF RF$_3$ and RF$_2$. The decoder then faces several candidates: RSF RF$_3$ spills beyond the hull, activates the upper left for the tank, and activates both banks for the dag fusion rather than the target. Finally, at RSF RF$_2$ the background returns to neutral blue, while at RSF RF$_1$ the response contracts onto the objects.

\section{Conclusion}

This paper proposes Cross-modal Progressive Filtering (CMPF) for referring remote sensing image segmentation. In contrast to existing multi-scale frameworks, which indiscriminately propagate cross-modally aligned features, CMPF progressively separates referring-relevant responses from irrelevant activations across scales. Specifically, cross-modal alignment is performed at each scale, Referring-aware Filtering (RF) is employed to suppress expression-irrelevant responses, and Selective Feature Injection (SFI) is used to selectively transfer informative features to subsequent stages. In addition, during decoding, the Referring-aware Selective Fusion (RSF) module adaptively aggregates multi-scale representations. Extensive experiments on three RRSIS benchmarks validate the effectiveness of CMPF, especially under strict IoU thresholds; moreover, ablation studies and response visualizations demonstrate that progressive filtering and selective propagation effectively alleviate distracting activations while enhancing target localization.

\bibliographystyle{IEEEtran}
\bibliography{ref}






%

\ifCLASSOPTIONcaptionsoff
  \newpage
\fi

\end{document}